\documentclass{article} 
\usepackage{iclr2026_conference,times}

\usepackage{amsmath,amsfonts,bm}

\def\eqref#1{equation~\ref{#1}}

\def\1{\bm{1}}

\DeclareMathAlphabet{\mathsfit}{\encodingdefault}{\sfdefault}{m}{sl}
\SetMathAlphabet{\mathsfit}{bold}{\encodingdefault}{\sfdefault}{bx}{n}

\usepackage{hyperref}
\usepackage{url}
\usepackage{booktabs}
\usepackage{graphicx}
\usepackage{subcaption}
\usepackage{pifont}
\usepackage{multirow}

\title{NGRPO: Negative-enhanced Group Relative Policy Optimization}
\iclrfinaltrue

\author{
Gongrui Nan, Siye Chen, Jing Huang, Mengyu Lu \\
\textbf{Dexun Wang, Chunmei Xie, Weiqi Xiong, Xianzhou Zeng} \\
\textbf{Qixuan Zhou, Yadong Li, Xingzhong Xu} \\
Ant Group \\
Shanghai, China \\
\texttt{\{nangongrui.ngr\}@antgroup.com}
}

\begin{document}

\maketitle

\begin{abstract}
RLVR has enhanced the reasoning capabilities of Large Language Models (LLMs) across various tasks. However, GRPO, a representative RLVR algorithm, suffers from a critical limitation: when all responses within a group are either entirely correct or entirely incorrect, the model fails to learn from these homogeneous responses. This is particularly problematic for homogeneously incorrect groups, where GRPO's advantage function yields a value of zero, leading to null gradients and the loss of valuable learning signals. To overcome this issue, we propose \textbf{NGRPO} (\textbf{N}egative-enhanced \textbf{G}roup \textbf{R}elative \textbf{P}olicy \textbf{O}ptimization), an algorithm designed to convert homogeneous errors into robust learning signals. First, NGRPO introduces \textbf{Advantage Calibration}. This mechanism hypothesizes the existence of a virtual maximum-reward sample during advantage calculation, thereby altering the mean and variance of rewards within a group and ensuring that the advantages for homogeneously incorrect samples are no longer zero. Second, NGRPO employs \textbf{Asymmetric Clipping}, which relaxes the update magnitude for positive samples while imposing stricter constraints on that of negative samples. This serves to stabilize the exploration pressure introduced by the advantage calibration. Our experiments on Qwen2.5-Math-7B demonstrate that NGRPO significantly outperforms baselines such as PPO, GRPO, DAPO, and PSR-NSR on mathematical benchmarks including MATH500, AMC23, and AIME2025. These results validate NGRPO's ability to learn from homogeneous errors, leading to stable and substantial improvements in mathematical reasoning. Our code is available at \url{https://github.com/nangongrui-ngr/NGRPO}.
\end{abstract}

\section{Introduction}
Large Reasoning Models (LRMs)~\citep{mathreasoning1,mathreasoning2,mathreasoning3} are renowned for their ability to reason about complex problems and have emerged as one of the most active research areas in artificial intelligence. Reinforcement Learning (RL) serves as an effective paradigm, playing an increasingly crucial role in enhancing the reasoning capabilities of LRMs. Among RL methods, algorithms like Proximal Policy Optimization (PPO)~\citep{ppo} are widely adopted for their stability and sample efficiency in complex, multi-step domains such as mathematical reasoning. The RLVR (Reinforcement Learning with Verifiable Rewards) technique~\citep{rlvr1,rlvr2,rlvr3,rlvr4,rlvr5} replaces the critic model in PPO with verifiable rewards, which simplifies the training process and enhances its stability. Group Relative Policy Optimization (GRPO)~\citep{grpo} is a representative RLVR algorithm. During training, GRPO generates multiple responses for a single prompt, which are then partitioned into a group. It determines the magnitude and direction of the policy optimization gradient by calculating and normalizing the rewards of the responses within the group. This dual capability of learning from both positive and negative samples has led to substantial improvements in the model's reasoning abilities.

Despite its impressive performance, GRPO suffers from a critical limitation: when a group is homogeneous, containing either all correct or all incorrect responses, the model fails to learn from it. This is because the variance of rewards within such a group is zero, which results in normalized advantages of zero and, consequently, a null gradient. As a result, the policy model misses valuable learning signals. This limitation is particularly detrimental for homogeneously incorrect groups, as the model is unable to learn from collective failures. Consequently, problems that the policy model failed to solve pre-training are unlikely to be solved post-training. In such cases, the policy model should be encouraged to engage in further exploration to find new solutions, rather than abandoning the problem.

PSR-NSR~\citep{psr_nsr} acknowledges the importance of negative samples, positing that learning from them increases the information entropy of the model's responses, where higher entropy signifies stronger exploration capabilities~\citep{nsr1,nsr2,nsr3}. PSR-NSR assigns a fixed negative advantage to negative samples and a fixed positive advantage to positive ones, aiming to enhance learning from negative samples and boost exploration by down-weighting the advantages of positive samples. However, this fixed-advantage assignment can lead to training collapse, particularly when faced with homogeneously negative samples~\citep{grpo_advantage_analyse} (as shown in Figure~\ref{fig:pnsr_collapse}). Therefore, a core challenge remains: how to effectively utilize these difficult negative samples to drive exploration and improve the model's reasoning capabilities, all while maintaining training stability.

To address this challenge, we introduce \textbf{NGRPO} (\textbf{N}egative-enhanced \textbf{G}roup \textbf{R}elative \textbf{P}olicy \textbf{O}ptimization), an algorithm designed to convert signals from homogeneous failures into robust exploration signals. NGRPO introduces an \textbf{Advantage Calibration} mechanism. By incorporating a virtual, maximum-reward sample into the standard GRPO advantage calculation, it ensures that advantages for homogeneously negative samples become negative. Without altering the gradient direction, this mechanism also reduces the gradient magnitude for positive samples while increasing it for negative ones, thereby further enhancing the model's exploration capabilities. Furthermore, we employ an \textbf{Asymmetric Clipping} mechanism. This mechanism stabilizes the exploration pressure introduced by Advantage Calibration by relaxing the update constraints on positive samples while imposing stricter ones on negative samples. Our main contributions are as follows:

\begin{enumerate}
\item We introduce the Advantage Calibration mechanism, which enables the model to learn from homogeneously negative samples and enhances its exploration capabilities.
\item We employ the Asymmetric Clipping mechanism, which works in concert with Advantage Calibration to stabilize the training process.
\item Experimental results on Qwen2.5-Math-7B~\citep{qwen2.5-math} demonstrate that our method significantly outperforms baselines such as PPO and GRPO on mathematical benchmarks, including MATH500~\citep{MATH}, AMC23~\citep{amc}, and AIME2025~\citep{aime}.
\end{enumerate}

\begin{figure}[htbp]
\centering
\includegraphics[width=\textwidth]{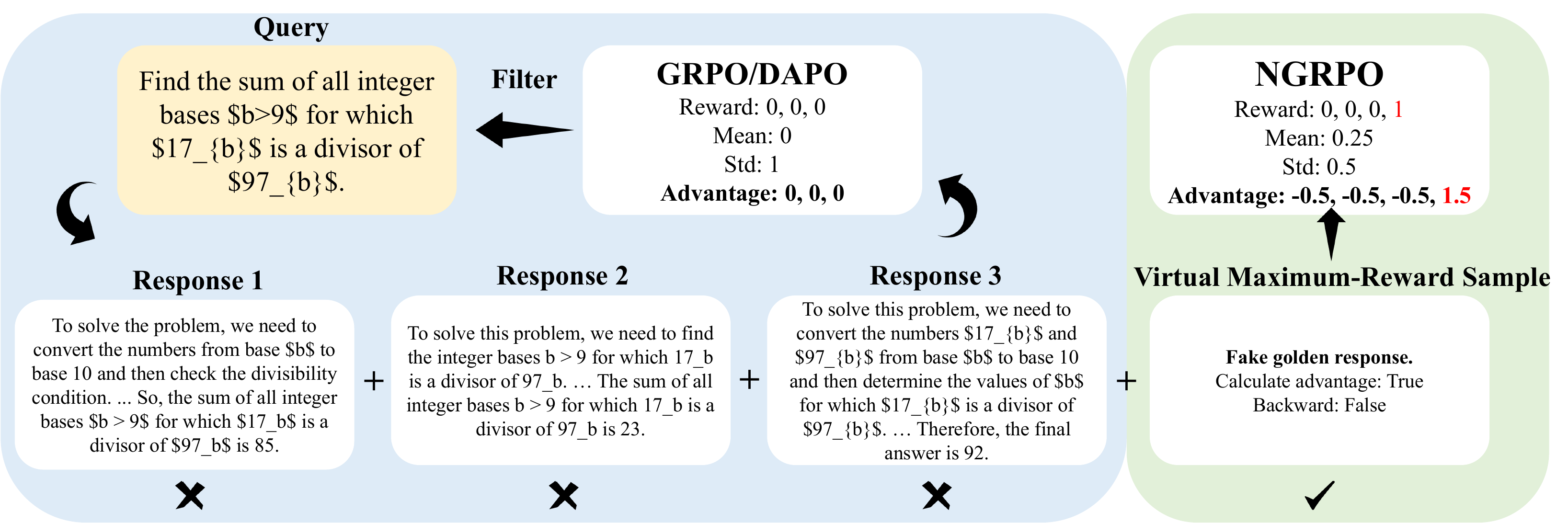}
\caption{Schematic of NGRPO's core mechanism. By introducing a virtual maximum-reward sample, NGRPO can generate learning signals even from homogeneously incorrect groups, turning failures into new opportunities for exploration.}
\label{fig:main}
\end{figure}

\section{Preliminaries}
GRPO (Group Relative Policy Optimization)~\citep{grpo}, much like PPO (Proximal Policy Optimization)~\citep{ppo}, is a policy gradient-based reinforcement learning algorithm, with GRPO building upon the foundation of PPO. Firstly, it dispenses with the critic model, employing verifiable rewards instead to assess the quality of a response. Secondly, it generates multiple responses for each prompt, which are assigned to a single group. By normalizing the rewards of responses within the same group, it computes a relative advantage for each response, which is then used to optimize the policy model. The objective function of GRPO is shown in Equation~\ref{equ:grpo}.

\begin{equation}
\label{equ:grpo}
\begin{split}
    \mathcal{J}_{G R P O}(\theta) = \mathbb{E}_{q \sim P(Q),\left\{o_{i}\right\}_{i = 1}^{G} \sim \pi_{\theta o l d}(o \mid q)}\left\{\frac { 1 } { G } \sum _ { i = 1 } ^ { G } \frac { 1 } { | o _ { i } | } \sum _ { t = 1 } ^ { | o _ { i } | } \left\{\operatorname { m i n } \left[\frac{\pi_{\theta}\left(o_{i, t} \mid q, o_{i,<t}\right)}{\pi_{\theta_{o l d}}\left(o_{i, t} \mid q, o_{i,<t}\right)} A_{i}, \right.\right.\right. \\
    \left.\left.\left.\operatorname{clip}\left(\frac{\pi_{\theta}\left(o_{i, t} \mid q, o_{i,<t}\right)}{\pi_{\theta_{o l d}}\left(o_{i, t} \mid q, o_{i,<t}\right)}, 1-\epsilon, 1+\epsilon\right) A_{i}\right]-\beta \mathbb{D}_{K L}\left[\pi_{\theta}| | \pi_{r e f}\right]\right\}\right\} ,
\end{split}
\end{equation}

\begin{equation}
    \mathbb{D}_{K L}\left[\pi_{\theta}| | \pi_{r e f}\right]=\frac{\pi_{r e f}\left(o_{i, t} \mid q, o_{i,<t}\right)}{\pi_{\theta}\left(o_{i, t} \mid q, o_{i,<t}\right)}-\log \frac{\pi_{r e f}\left(o_{i, t} \mid q, o_{i,<t}\right)}{\pi_{\theta}\left(o_{i, t} \mid q, o_{i,<t}\right)}-1 ,
\end{equation}

\begin{equation}
    A_{i}=\frac{r_{i}-\operatorname{mean}\left(\left\{r_{1}, r_{2}, \ldots, r_{G}\right\}\right)}{\operatorname{std}\left(\left\{r_{1}, r_{2}, \ldots, r_{G}\right\}\right)} .
\end{equation}

Here, $\pi_{\theta}$ represents the current policy model, and $\pi_{\theta_{\text{old}}}$ ($\pi_{\theta_{\text{ref}}}$) represents the old (reference) policy model. $D_{KL}$ denotes the KL divergence between the current and reference policy models. It ensures that the policy model does not deviate too far from the reference model during updates. The term $A$ represents the intra-group advantage. Its calculation depends on the mean and variance of the rewards $r_i$, for each response within the group. Within a group, responses with rewards higher than the average receive a positive advantage, while those with lower rewards receive a negative advantage.

When encountering a homogeneous group, where all rewards are identical—the mean reward equals each individual reward, causing the advantage for every sample in that group to be zero. Although GRPO's reward normalization numerically stabilizes the training process, it diminishes the utilization of training samples, particularly for datasets with either extremely high or extremely low difficulty. DAPO (Dynamic Sampling Policy Optimization)~\citep{dapo} employs a pre-filtering mechanism that discards both homogeneously correct and homogeneously incorrect samples, preventing them from participating in the policy model's optimization. DAPO improves training efficiency and avoids the issue of high gradient variance caused by sparse advantages, thereby further stabilizing the training process and enhancing the model's capabilities. However, it still fails to fully utilize the available data, particularly the homogeneously incorrect samples.

\section{Motivation}
As analyzed in~\citep{dose_rl_works}, reinforcement learning may not be able to impart new capabilities to a model beyond what it acquired during pre-training. Consequently, LUFFY~\citep{luffy} replaces one response in a group with an output from a more powerful model, using the new reasoning trajectory to instruct the model under training. Similarly, RELIFT~\citep{relift} imparts new knowledge to the model through a seamless transition between RL and SFT. However, the recent work of PSR-NSR~\citep{psr_nsr} demonstrates the importance and potential of encouraging model exploration. It shows that even without introducing new models or responses, a model can achieve significant performance gains solely through reinforcement learning, a benefit attributed to learning from negative samples.

To ensure that samples in a homogeneous group have non-zero advantages, a natural approach is to dispense with the normalization process of the original GRPO. This is precisely the strategy adopted by PSR-NSR, which, for mathematical tasks, assigns a fixed advantage of -1 to negative samples and 0.1 to positive samples, as shown in Equation~\ref{equ:fixed_advantage}.
\begin{equation}
\label{equ:fixed_advantage}
A_{i} =
\begin{cases}
0.1, & y_{i} \text{ is equal to } y \\
-1, & y_{i} \text{ is not equal to } y
\end{cases}
\end{equation}
where $y_{i}$ is the model's outputted answer and $y$ is the correct answer.

\begin{figure}[htbp]
\centering
\begin{subfigure}[t]{0.48\textwidth}
\centering
\includegraphics[width=\linewidth]{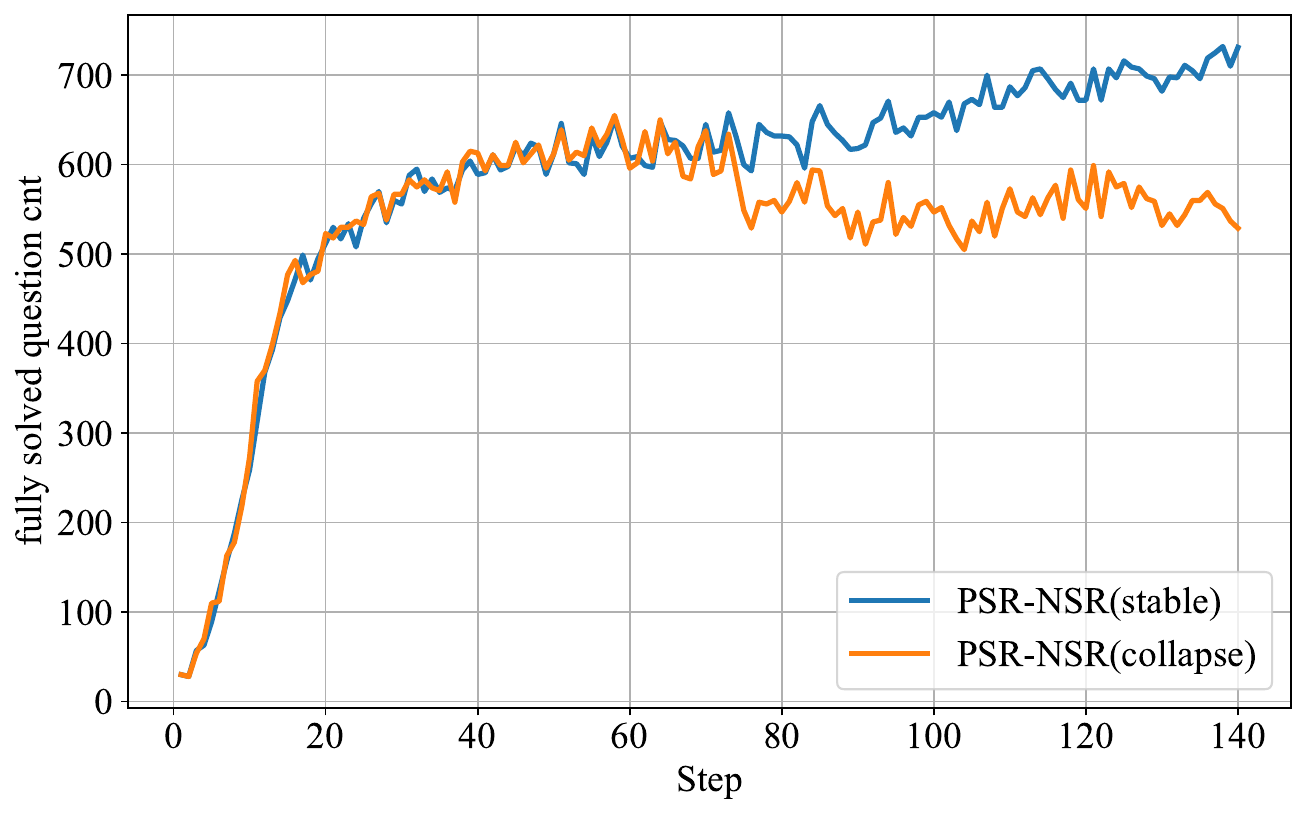}
\caption{Fully solved questions.}
\end{subfigure}
\hfill
\begin{subfigure}[t]{0.48\textwidth}
\centering
\includegraphics[width=\linewidth]{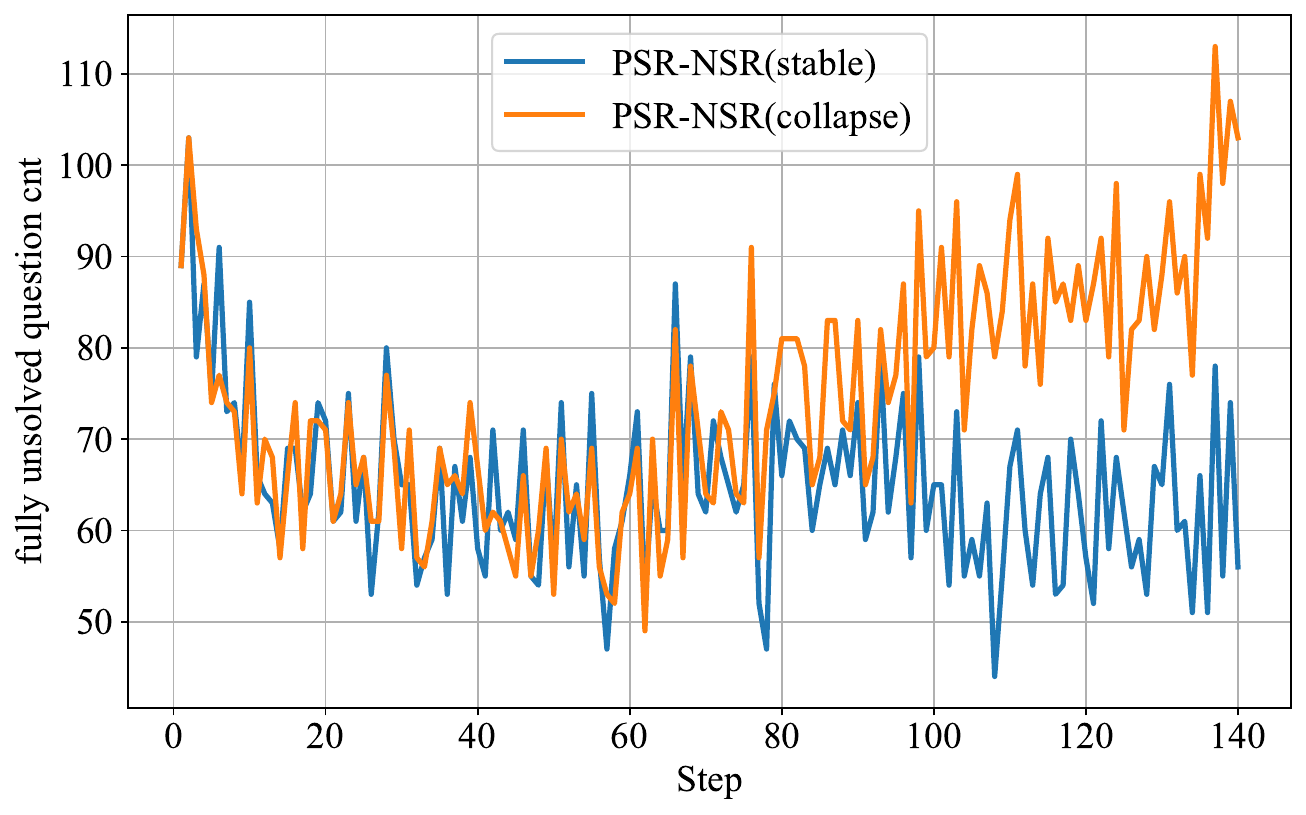}
\caption{Fully unsolved questions.}
\end{subfigure}
\vspace{1em}
\begin{subfigure}[t]{0.48\textwidth}
\centering
\includegraphics[width=\linewidth]{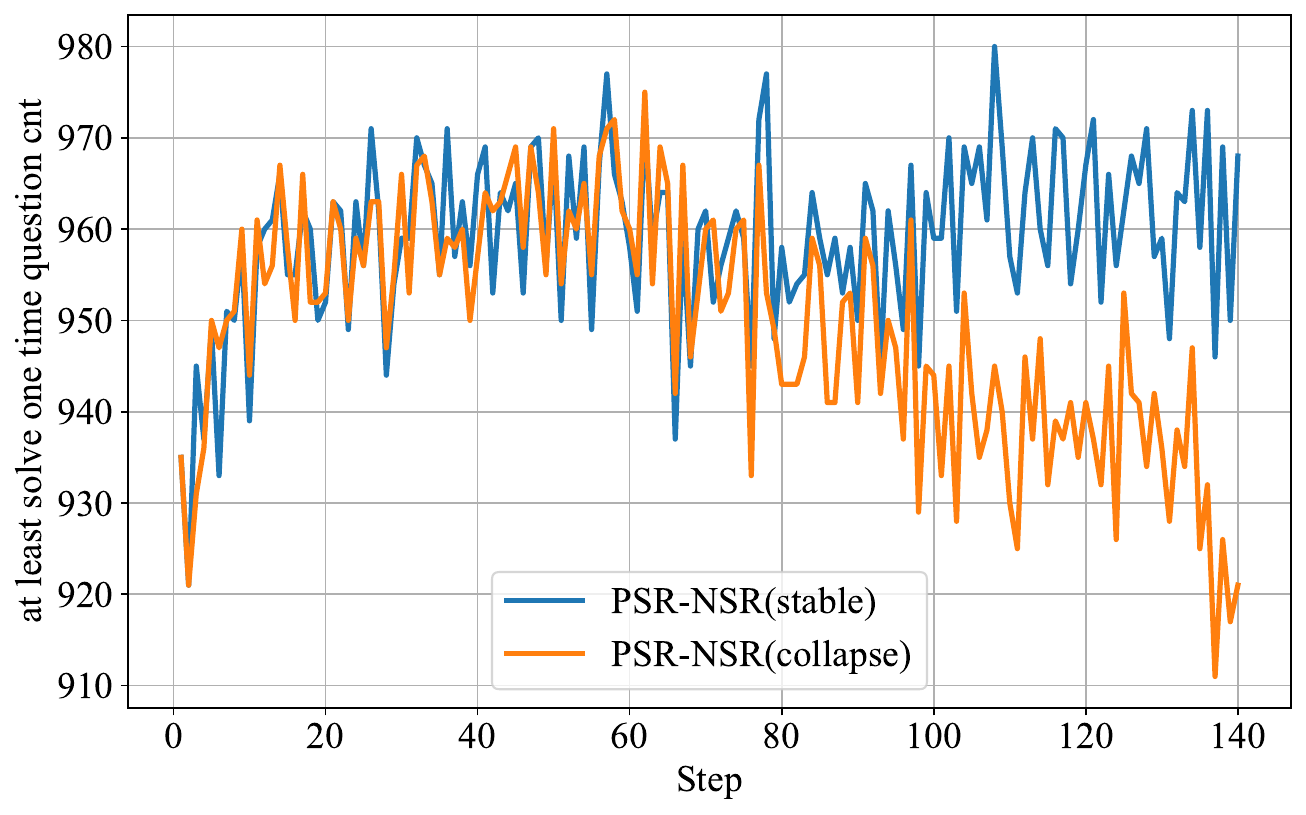}
\caption{Questions solved at least once.}
\end{subfigure}
\hfill
\begin{subfigure}[t]{0.48\textwidth}
\centering
\includegraphics[width=\linewidth]{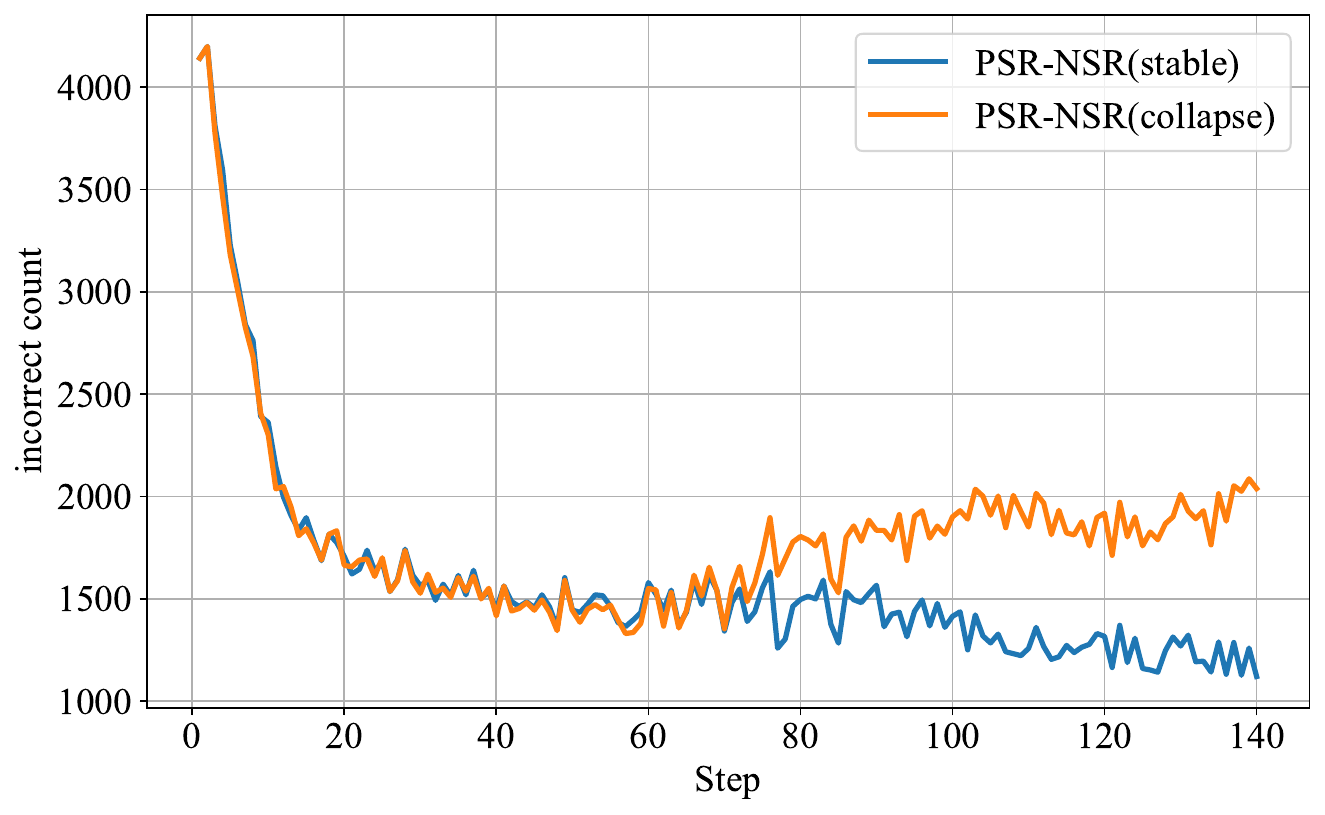}
\caption{Total incorrect responses.}
\end{subfigure}
\caption{The fixed negative advantage in PSR-NSR can easily lead to training collapse. The figure compares a stable run (blue) with a collapsed run (orange). The collapse is characterized by a sharp decline in the number of fully solved questions and a simultaneous surge in the number of incorrect responses. This highlights the risks of training on homogeneously incorrect groups without advantage normalization.}
\label{fig:pnsr_collapse}
\end{figure}

However, we observe that the stability of model training is highly sensitive to the values of these fixed advantages. Specifically, a larger magnitude for the negative advantage, relative to the positive one, leads to greater training instability, as illustrated in Figure~\ref{fig:pnsr_collapse}. Conversely, reducing the magnitude of the negative advantage suppresses the model's exploration capabilities. Consequently, this method struggles to strike a trade-off between exploration and training stability.

Similarly, experiments in~\citep{grpo_advantage_analyse} corroborate this finding. For the REINFORCE~\citep{reinforce} algorithm, applying either advantage normalization or filtering out homogeneous negative samples allows its performance to approach that of GRPO. However, if homogeneous negative samples are retained without advantage normalization, the algorithm's performance degrades severely. In other words, enabling the model to learn from homogeneously incorrect groups seems to necessitate advantage normalization to prevent the training process from collapsing.

Therefore, our objective is to devise a method that enables learning from homogeneously incorrect groups while retaining the stability benefits of advantage normalization.

\section{Negative-enhanced Group Relative Policy Optimization (NGRPO)}
To address the limitations of GRPO, we introduce Negative-enhanced Group Relative Policy Optimization (NGRPO), an extension designed to effectively learn from failure and promote robust exploration, as illustrated in Figure~\ref{fig:main}.

\subsection{Advantage Calibration}
To resolve the zero-advantage issue in homogeneous-incorrect groups, NGRPO introduces a \textbf{Virtual Maximum-Reward Sample}. The core idea is to augment the advantage calculation for every group by conceptually adding a sample that achieved the maximum possible reward, $r_{\text{max}}$. This virtual sample is not generated by the policy; only its reward value is used to enrich the normalization statistics.

Let $\mathcal{R} = \{r_1, \ldots, r_G\}$ be the set of rewards from the $G$ sampled responses. We define an augmented reward set $\mathcal{R}' = \mathcal{R} \cup \{r_{\text{max}}\}$. The NGRPO advantage, $A'_i$, is then calculated by standardizing $r_i$ over this augmented set:
\begin{align}
A'_{i} = \frac{r_{i} - \mu'_{\mathcal{R}}}{\sigma'_{\mathcal{R}} + \epsilon_{\text{std}}},
\end{align}
where $\mu'_{\mathcal{R}}$ and $\sigma'_{\mathcal{R}}$ are the mean and standard deviation of the augmented rewards:
\begin{align}
\mu'_{\mathcal{R}} = \frac{1}{G+1} \left( \left( \sum_{j=1}^G r_j \right) + r_{\text{max}} \right), \quad \sigma'_{\mathcal{R}} = \sqrt{\frac{1}{G+1}\sum_{r \in \mathcal{R}'} (r - \mu'_{\mathcal{R}})^2}.
\end{align}

The objective function of NGRPO is shown in Equation~\ref{equ:ngrpo}.
\begin{equation}
\label{equ:ngrpo}
\begin{split}
    \mathcal{J}_{N G R P O}(\theta) = \mathbb{E}_{q \sim P(Q),\left\{o_{i}\right\}_{i = 1}^{G} \sim \pi_{\theta o l d}(o \mid q)}\left\{\frac { 1 } { G } \sum _ { i = 1 } ^ { G } \frac { 1 } { | o _ { i } | } \sum _ { t = 1 } ^ { | o _ { i } | } \left\{\operatorname { m i n } \left[\frac{\pi_{\theta}\left(o_{i, t} \mid q, o_{i,<t}\right)}{\pi_{\theta_{o l d}}\left(o_{i, t} \mid q, o_{i,<t}\right)} A'_{i}, \right.\right.\right. \\
    \left.\left.\left.\operatorname{clip}\left(\frac{\pi_{\theta}\left(o_{i, t} \mid q, o_{i,<t}\right)}{\pi_{\theta_{o l d}}\left(o_{i, t} \mid q, o_{i,<t}\right)}, 1-\epsilon, 1+\epsilon\right) A'_{i}\right]\right\}\right\} ,
\end{split}
\end{equation}

Conceptually, introducing $r_{\text{max}}$ creates a dynamic, batch-dependent baseline for advantage calculation that is always above the observed rewards. This guarantees a meaningful negative advantage in homogeneous-incorrect groups, where GRPO would stagnate. Furthermore, the mechanism is inherently adaptive: its impact on both the mean and the standard deviation applies a stronger push for exploration when group performance is uniformly low, while minimally affecting high-performing groups. This approach preserves training stability by modulating the magnitude of advantages to encourage exploration without altering their fundamental sign.

\begin{figure}[htbp]
\centering
\includegraphics[width=\textwidth]{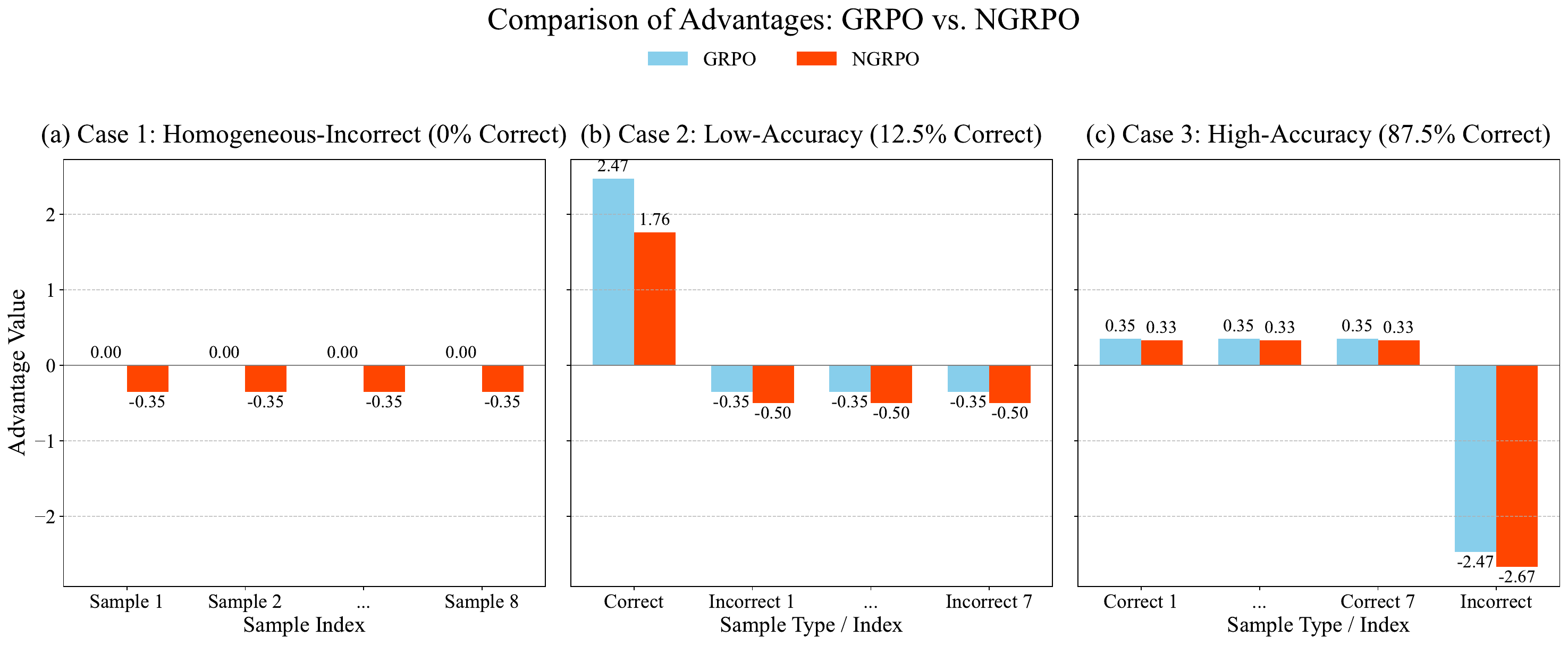}
\caption{Comparison of advantage distributions for GRPO and NGRPO across three representative scenarios with a group size of $G=8$. NGRPO generates learning signals from failures and adaptively re-balances advantages to promote exploration.}
\label{fig:case}
\end{figure}

The adaptive behavior of NGRPO is illustrated across three representative cases in Figure~\ref{fig:case}:
\begin{itemize}
    \item \textbf{Case 1: Homogeneous-Incorrect Group (0\% correct).} GRPO provides no learning signal (zero advantage). In contrast, NGRPO generates a uniform negative advantage, enabling the model to learn even from complete failure.
    \item \textbf{Case 2: Low-Accuracy Mixed Group (12.5\% correct).} NGRPO significantly re-balances the advantages. It dampens the advantage for the correct sample (from 2.47 to 1.76) while increasing the penalty for incorrect ones (from -0.35 to -0.50). This strong adjustment promotes exploration when model performance is low.
    \item \textbf{Case 3: High-Accuracy Mixed Group (87.5\% correct).} In this high-accuracy scenario, NGRPO's adjustments are minimal, slightly reducing positive advantages while moderately increasing the penalty for the single failure. This demonstrates the method's adaptability, favoring exploitation when performance is already strong.
\end{itemize}

\subsection{Asymmetric Clipping}
In standard GRPO, the sum of advantages across a group is zero ($\sum_{i=1}^G A_i = 0$). The introduction of $r_{\text{max}}$, however, ensures the sum of NGRPO advantages is always negative ($\sum_{i=1}^G A'_i < 0$). This creates a persistent negative bias in the policy update, effectively increasing policy entropy and encouraging exploration. While beneficial, this strong exploratory pressure can increase the risk of training instability.

To counteract this, we use \textbf{Asymmetric Clipping} to adjust the PPO clipping range. Instead of the standard symmetric range $[1-\epsilon, 1+\epsilon]$, we define separate clipping boundaries for positive and negative advantages. The modified clipped objective is:
\begin{equation}
L_t^{\text{N-CLIP}}(\theta) = 
\begin{cases} 
\min(\rho_t(\theta) A'_i, (1 + \epsilon_{\text{pos}}) A'_i) & \text{if } A'_i \geq 0 \\
\max(\rho_t(\theta) A'_i, (1 - \epsilon_{\text{neg}}) A'_i) & \text{if } A'_i < 0 
\end{cases}
\end{equation}
where $\epsilon_{\text{pos}}$ and $\epsilon_{\text{neg}}$ are distinct hyperparameters. To stabilize the strong exploratory drive from our virtual sample, we set a tighter constraint on negative updates ($\epsilon_{\text{neg}}=0.16$) while allowing more latitude for positive updates ($\epsilon_{\text{pos}}=0.24$), following \citet{dapo}. This asymmetric design is crucial for balancing the strong, persistent exploratory pressure introduced by our virtual sample. By penalizing incorrect paths more cautiously while strongly rewarding correct ones, NGRPO can harness the benefits of enhanced exploration without compromising the training stability essential for effective learning.

\section{Experiments and Results}
\subsection{Experimental Setup}
\subsubsection{Training and Evaluation}
All models were trained for 20 epochs on the MATH dataset~\citep{MATH} using a cluster of 8 NVIDIA H100 GPUs. Key hyperparameters included a learning rate of 1e-6, a global batch size of 1024, a maximum prompt length of 1024, and a maximum response length of 3072 for the Qwen2.5-Math-7B model~\citep{qwen2.5-math}. Models were subsequently evaluated on three mathematical reasoning benchmarks: MATH500, AMC23~\citep{amc}, and the highly challenging AIME2025~\citep{aime}.

\subsubsection{Metrics}
We evaluated all models using the Unbiased Pass@$k$ metric~\citep{unbiased-pass@k}, generating $N=256$ samples per problem with a temperature of 0.6 and top-p of 0.95. To provide a single, holistic measure of performance, we report the Pass@$k$ Area Under the Curve (AUC). This metric aggregates performance across all $k$ values ($k \in \{1, 2, \dots, 256\}$) by applying the trapezoidal rule to the Pass@$k$ curve on a log-log scale, offering a comprehensive view of a model's overall problem-solving capability.

\subsubsection{Baselines}
We benchmarked NGRPO against several prominent RL algorithms: Proximal Policy Optimization (PPO)~\citep{ppo}, Group Relative Policy Optimization (GRPO)~\citep{grpo}, Dispersed Advantage Policy Optimization (DAPO)~\citep{dapo}, and PSR-NSR~\citep{psr_nsr}.

\subsection{Comparative Analysis}
\label{sec:qwen2.5-7b}
\begin{figure}[htbp]
\centering
\includegraphics[width=\textwidth]{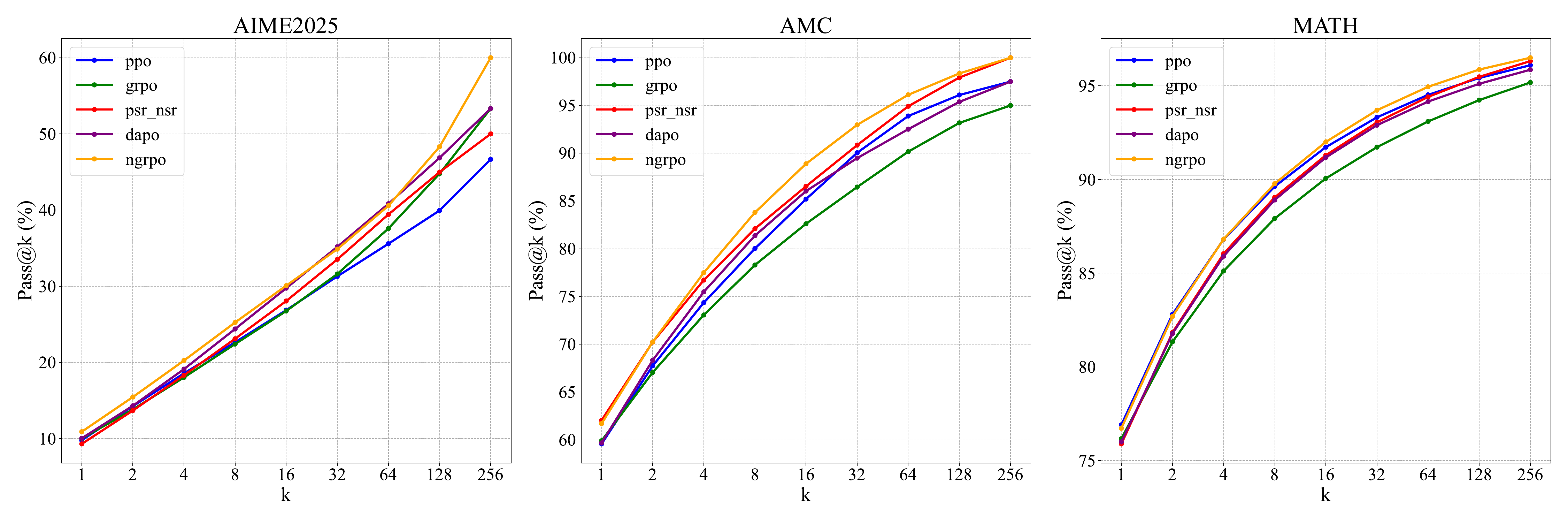}
\caption{Pass@$k$ performance curves for different RL methods on AIME2025, AMC, and MATH test sets, using the Qwen2.5-Math-7B model. NGRPO consistently outperforms or matches the strongest baselines across varying $k$ values.}
\label{fig:qwen2.5-7b}
\end{figure}

As shown in Table~\ref{tab:qwen2.5-7b} and Figure~\ref{fig:qwen2.5-7b}, NGRPO demonstrates superior performance across the board. On the highly challenging AIME2025 dataset, NGRPO achieves the highest Pass@$k$ AUC, indicating a strong and balanced improvement in both low-$k$ accuracy and high-$k$ exploration. This result strongly supports our central hypothesis: enabling a model to learn from collective failure is critical for unlocking new reasoning capabilities, especially on problems far beyond its initial grasp. In contrast, methods like GRPO and DAPO, which filter out homogeneous-incorrect groups, appear to hit a performance ceiling, likely due to their limited exposure to difficult problem variations. NGRPO's consistent state-of-the-art results underscore the effectiveness of transforming failure into a productive learning signal.

\begin{table}[htbp]
\centering
\caption{Performance comparison of various RL methods on the Qwen2.5-Math-7B model. Results are reported as percentages (\%). The best results are in \textbf{bold}, and the second-best are \underline{underlined}.}
\label{tab:qwen2.5-7b}
\resizebox{\textwidth}{!}{
\begin{tabular}{lcccccccccc}
\toprule
\multirow{2}{*}{\textbf{Method}} & \multicolumn{9}{c}{\textbf{Pass@$k$}} & \textbf{Pass@$k$} \\
\cmidrule(lr){2-10}
 & $k=1$ & $k=2$ & $k=4$ & $k=8$ & $k=16$ & $k=32$ & $k=64$ & $k=128$ & $k=256$ & \textbf{AUC} \\
\midrule
\multicolumn{11}{c}{\textbf{AIME2025}} \\
\midrule
PPO & 9.78 & 14.25 & 18.57 & 22.65 & 26.86 & 31.30 & 35.58 & 39.94 & 46.67 & 27.17 \\
GRPO & 9.99 & 13.83 & 18.03 & 22.41 & 26.75 & 31.61 & 37.58 & 44.78 & \underline{53.33} & 28.33 \\
PSR-NSR & 9.30 & 13.68 & 18.32 & 23.11 & 28.06 & 33.53 & 39.43 & 44.98 & 50.00 & 28.85 \\
DAPO & \underline{10.04} & \underline{14.31} & \underline{19.12} & \underline{24.39} & \underline{29.75} & \textbf{35.18} & \textbf{40.83} & \underline{46.86} & \underline{53.33} & \underline{30.27} \\
NGRPO (Ours) & \textbf{10.90} & \textbf{15.46} & \textbf{20.25} & \textbf{25.25} & \textbf{30.08} & \underline{34.87} & \underline{40.57} & \textbf{48.31} & \textbf{60.00} & \textbf{31.28} \\
\midrule
\multicolumn{11}{c}{\textbf{AMC}} \\
\midrule
PPO & 59.57 & 67.75 & 74.36 & 80.03 & 85.20 & 90.05 & 93.90 & 96.10 & 97.50 & 83.24 \\
GRPO & 59.90 & 67.06 & 73.07 & 78.30 & 82.62 & 86.46 & 90.17 & 93.18 & 95.00 & 81.04 \\
PSR-NSR & \textbf{62.05} & \underline{70.23} & \underline{76.72} & \underline{82.10} & \underline{86.54} & \underline{90.85} & \underline{94.92} & \underline{97.93} & \textbf{100.00} & \underline{85.04} \\
DAPO & 59.67 & 68.33 & 75.49 & 81.37 & 86.04 & 89.49 & 92.52 & 95.38 & 97.50 & 83.40 \\
NGRPO (Ours) & \underline{61.69} & \textbf{70.24} & \textbf{77.50} & \textbf{83.80} & \textbf{88.90} & \textbf{92.97} & \textbf{96.12} & \textbf{98.36} & \textbf{100.00} & \textbf{86.09} \\
\midrule
\multicolumn{11}{c}{\textbf{MATH}} \\
\midrule
PPO & \textbf{76.91} & \textbf{82.81} & \underline{86.81} & \underline{89.63} & \underline{91.73} & \underline{93.32} & \underline{94.52} & 95.42 & 96.10 & \underline{90.09} \\
GRPO & 76.17 & 81.34 & 85.12 & 87.92 & 90.06 & 91.73 & 93.10 & 94.24 & 95.18 & 88.65 \\
PSR-NSR & 75.89 & 81.84 & 86.04 & 89.05 & 91.30 & 93.05 & 94.41 & \underline{95.48} & \underline{96.32} & 89.66 \\
DAPO & 76.00 & 81.78 & 85.91 & 88.91 & 91.18 & 92.90 & 94.16 & 95.11 & 95.86 & 89.49 \\
NGRPO (Ours) & \underline{76.73} & \underline{82.70} & \textbf{86.82} & \textbf{89.78} & \textbf{92.01} & \textbf{93.70} & \textbf{94.95} & \textbf{95.87} & \textbf{96.50} & \textbf{90.31} \\
\bottomrule
\end{tabular}}
\end{table}

\begin{figure}[htbp]
\begin{center}
\includegraphics[width=0.6\textwidth]{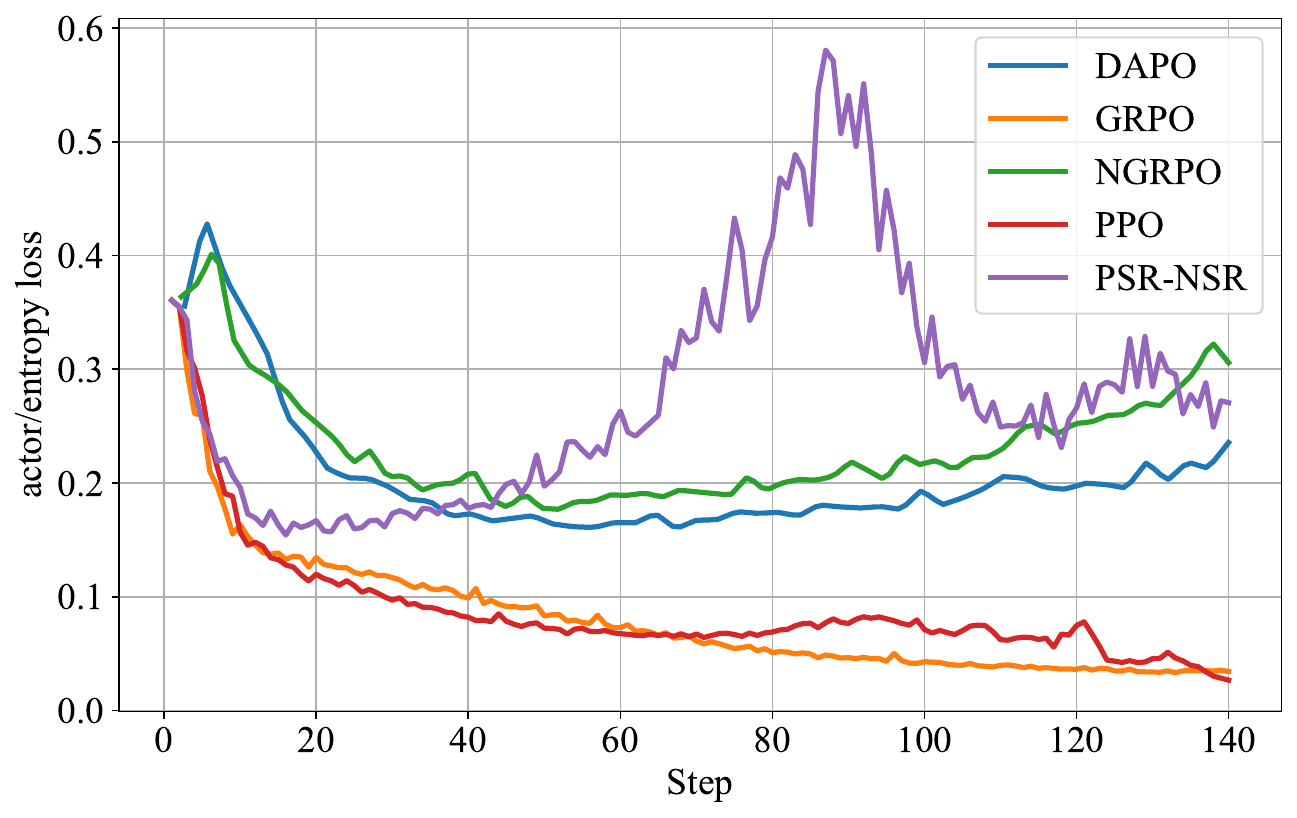}
\end{center}
\caption{Entropy changes during training. NGRPO achieves a steady increase after entropy decreases.}
\label{fig:entropy}
\end{figure}

Figure~\ref{fig:entropy} illustrates the policy entropy dynamics\citep{entropy}. NGRPO maintains a healthy, stable entropy that gradually converges, indicating a robust balance between exploration and exploitation. This contrasts sharply with DAPO, GRPO, and PPO, whose entropies continuously decline, signaling a progressive loss of exploratory capacity. Conversely, PSR-NSR maintains a significantly higher entropy, reflecting an aggressive and often unstable exploratory pressure, which aligns with the performance collapse shown in Figure~\ref{fig:pnsr_collapse}. NGRPO's entropy profile is thus consistent with its design for stable, adaptive exploration.

\subsection{Ablation Studies}
To dissect the contributions of NGRPO's key components, we conducted a series of ablation studies, with results summarized in Table~\ref{tab:ablation_components}. The baseline GRPO, which lacks all three of our proposed modifications, establishes the performance floor. Introducing only the asymmetric clipping range yields marginal gains, while adding the virtual maximum reward by itself provides a more substantial boost, confirming its effectiveness as a standalone enhancement. Crucially, the combination of asymmetric clipping and the virtual reward demonstrates a strong synergistic effect, outperforming the sum of their individual contributions. The full NGRPO model—which also retains homogeneous-incorrect groups—achieves the best performance, validating our claim that learning from these challenging samples is essential for maximizing model capability.

\begin{table}[htbp]
\centering
\caption{Ablation study of NGRPO components on Qwen2.5-Math-7B. The baseline (\ding{55}\ding{55}\ding{55}) corresponds to the standard GRPO algorithm.}
\label{tab:ablation_components}
\small
\begin{tabular}{ccccc}
\toprule
\multirow{2}{*}{\textbf{Advantage Calibration}} & \multirow{2}{*}{\textbf{Learning from failures}} & \multirow{2}{*}{\textbf{Asymmetric Clipping}} & \multicolumn{2}{c}{\textbf{Pass@$k$ AUC}} \\
\cmidrule(lr){4-5}
& & & \textbf{AIME2025} & \textbf{AMC} \\
\midrule
\ding{55} & \ding{55} & \ding{55} & 28.33 & 81.04 \\ 
\ding{55} & \ding{55} & \ding{51} & 28.48 & 81.22 \\
\ding{51} & \ding{55} & \ding{55} & 29.56 & 83.85 \\
\ding{51} & \ding{55} & \ding{51} & \underline{30.54} & \underline{84.37} \\
\ding{51} & \ding{51} & \ding{51} & \textbf{31.28} & \textbf{86.09} \\ 
\bottomrule
\end{tabular}
\end{table}

\section{Related Work}
\subsection{Reinforcement Learning for LLM Alignment}
Reinforcement Learning (RL) has become a cornerstone for aligning Large Language Models (LLMs) with human preferences and task objectives. 

Proximal Policy Optimization (PPO)~\citep{ppo}, in particular, has emerged as a de facto standard due to its stability and ease of implementation, achieving notable success in complex tasks like code generation and mathematical reasoning. While alternative paradigms like Direct Preference Optimization (DPO)~\citep{dpo} have gained traction by bypassing explicit reward modeling, PPO and its variants remain a critical tool for capability enhancement, especially in domains like mathematical reasoning that require extensive exploration to discover correct solutions. Our work builds upon the PPO framework, aiming to enhance its efficiency and robustness in challenging, low-reward scenarios.

\subsection{RLVR}
A significant challenge in applying PPO to LLMs is training a stable and accurate critic network to estimate the value function, which is notoriously difficult in the vast state-space of language. To address this, a line of critic-less RL algorithms has emerged. Group Relative Policy Optimization (GRPO)~\citep{grpo} is a prime example, eliminating the critic network by calculating advantages based on the relative ranking of rewards within a sampled group. This design significantly simplifies the training pipeline. However, as we have analyzed, GRPO's efficacy is contingent on reward variance. In homogeneous groups, where all samples receive the same reward, the advantage function collapses to zero, stalling the learning process and exposing a fundamental limitation of this otherwise efficient approach.

\subsection{Learning from Negative Samples}
Effectively utilizing negative samples (i.e., incorrect responses) is a core challenge in RL for reasoning. Existing methods have addressed GRPO's stagnation problem with varying trade-offs. DAPO~\citep{dapo} proposed filtering out homogeneous groups—a strategy of avoidance that discards valuable learning opportunities. In contrast, PSR-NSR~\citep{psr_nsr} adopted a more direct approach, assigning a fixed negative advantage to all incorrect responses. While demonstrating the potential of negative signals, this hard-coded penalty lacks normalization and can destabilize training, leading to performance collapse.

NGRPO offers a more principled solution. It neither discards negative samples like DAPO nor introduces an un-normalized penalty like PSR-NSR. By introducing a virtual maximum-reward sample, NGRPO generates a meaningful, adaptive, and non-zero advantage for homogeneous-incorrect groups. Paired with an asymmetric clipping mechanism for stability, our method transforms failure into a robust and effective learning signal. This places NGRPO as a more robust and stable paradigm for learning from failure within the GRPO framework.

\section{Conclusion}
We proposed NGRPO. \textbf{Advantage Calibration} addresses the issue of homogeneous error groups being unable to participate in training in GRPO by introducing a virtual maximum reward example, while also improving the model's exploration capabilities. To accommodate this enhanced exploration capability, we employed \textbf{Asymmetric Clipping}, allowing more freedom for the gradients of positive examples while constraining the gradients of negative examples to ensure training stability. Extensive experiments demonstrated the effectiveness of our proposed method. 

We will continue to explore the boundaries of model capabilities and validate our proposed method on more models and datasets.



\bibliographystyle{iclr2026_conference}

\appendix
\section{Appendix}

\subsection{Training Process Analysis}
To provide deeper insight into the training dynamics of NGRPO compared to baseline methods, we analyze several key metrics throughout the training process.

\subsubsection{Response Length}
\begin{figure}[htbp]
\begin{center}
    \begin{subfigure}[t]{0.48\textwidth}
    \centering
        \includegraphics[width=\linewidth]{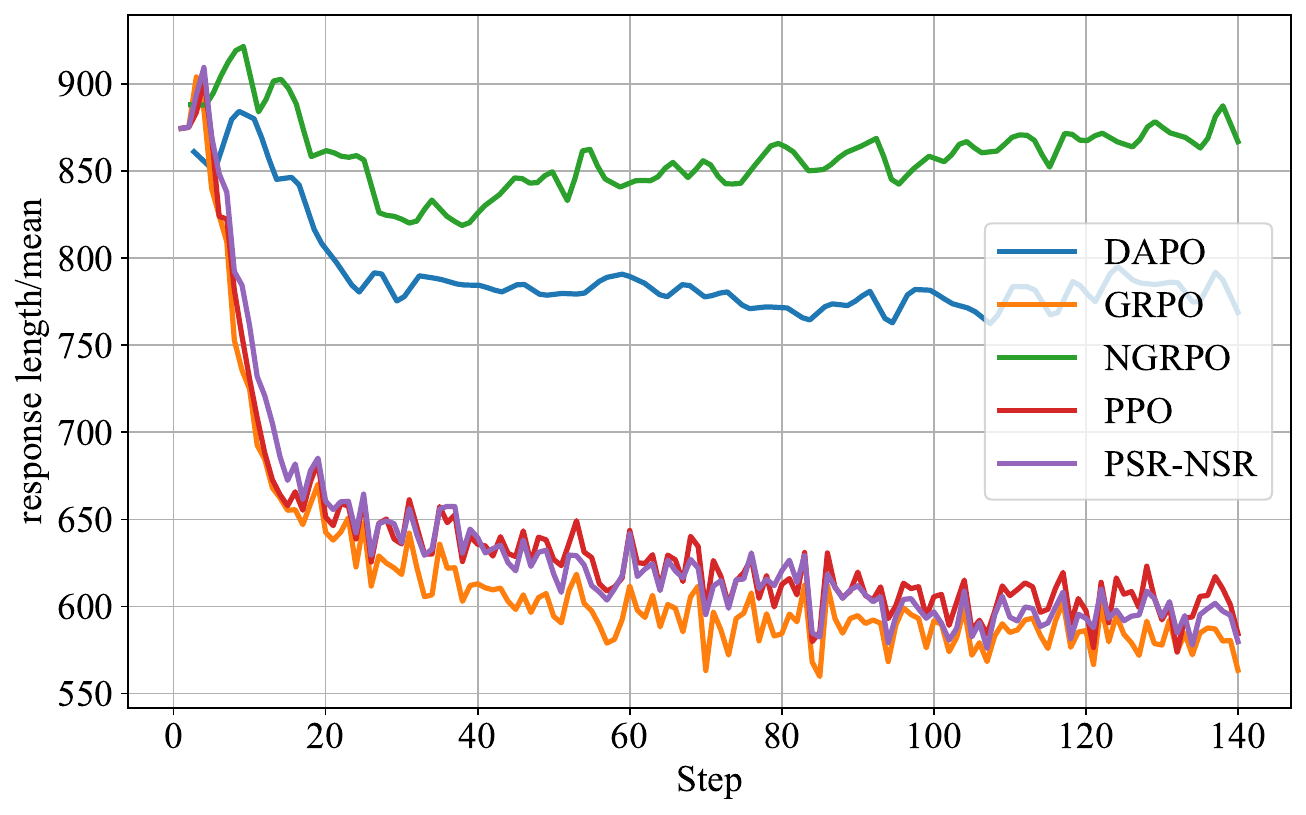}
    \caption{Mean response length during training.}
    \end{subfigure}
    \hfill
    \begin{subfigure}[t]{0.48\textwidth}
    \centering
        \includegraphics[width=\linewidth]{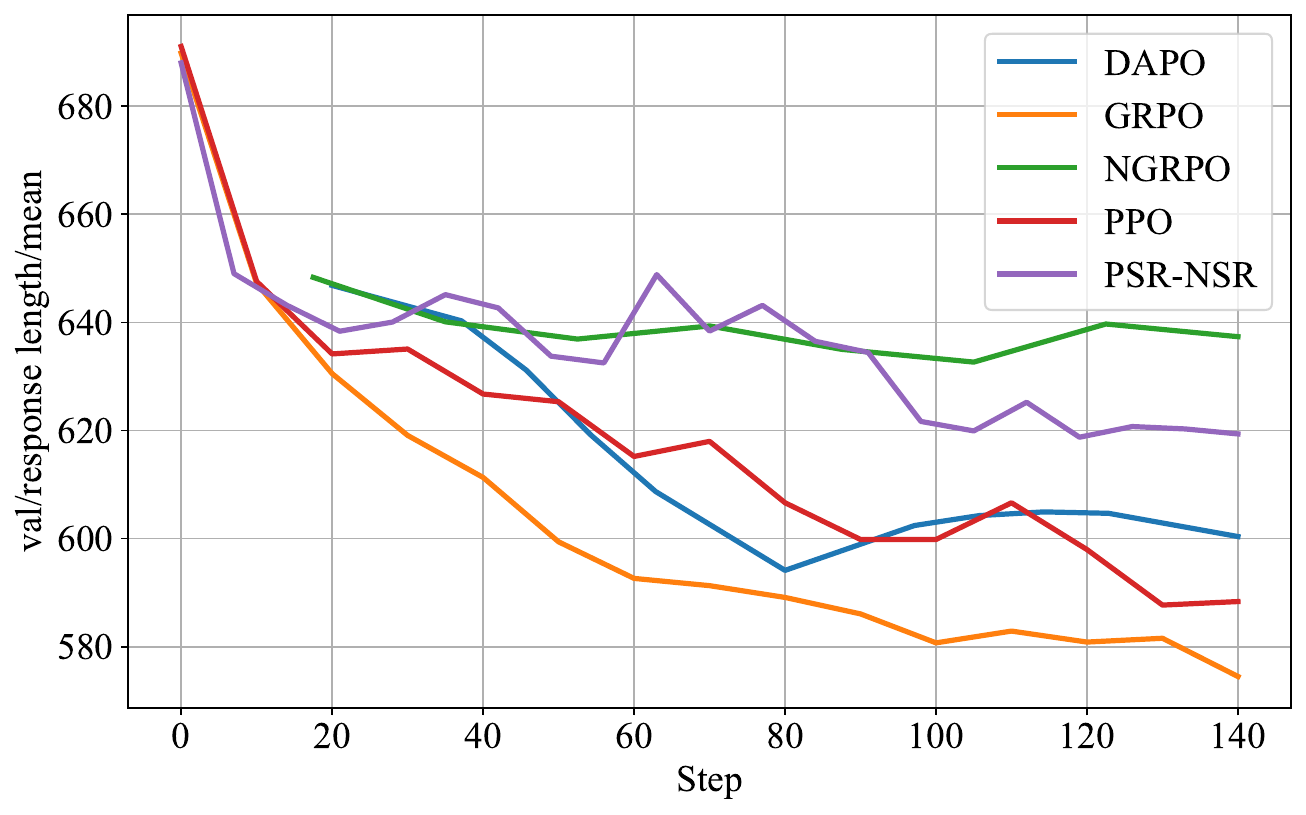}
    \caption{Mean response length during evaluation.}
    \end{subfigure}
\caption{Response length dynamics during training for different reinforcement learning methods.}
\label{fig:resp_length}
\end{center}
\end{figure}

Figure~\ref{fig:resp_length} shows that during training (left), NGRPO and DAPO generate longer responses because filtering homogeneous-correct groups forces them to focus on more challenging problems requiring longer reasoning chains. During evaluation (right), response length correlates with policy entropy. PSR-NSR produces the longest responses, while the lengths for DAPO, GRPO, and PPO steadily shrink. NGRPO's ability to first decrease and then stabilize its response length suggests it avoids collapsing into overly simplistic strategies, retaining its capacity for complex reasoning.

\subsubsection{Reward and Advantage}
\begin{figure}[htbp]
\begin{center}
    \begin{subfigure}[t]{0.48\textwidth}
    \centering
        \includegraphics[width=\linewidth]{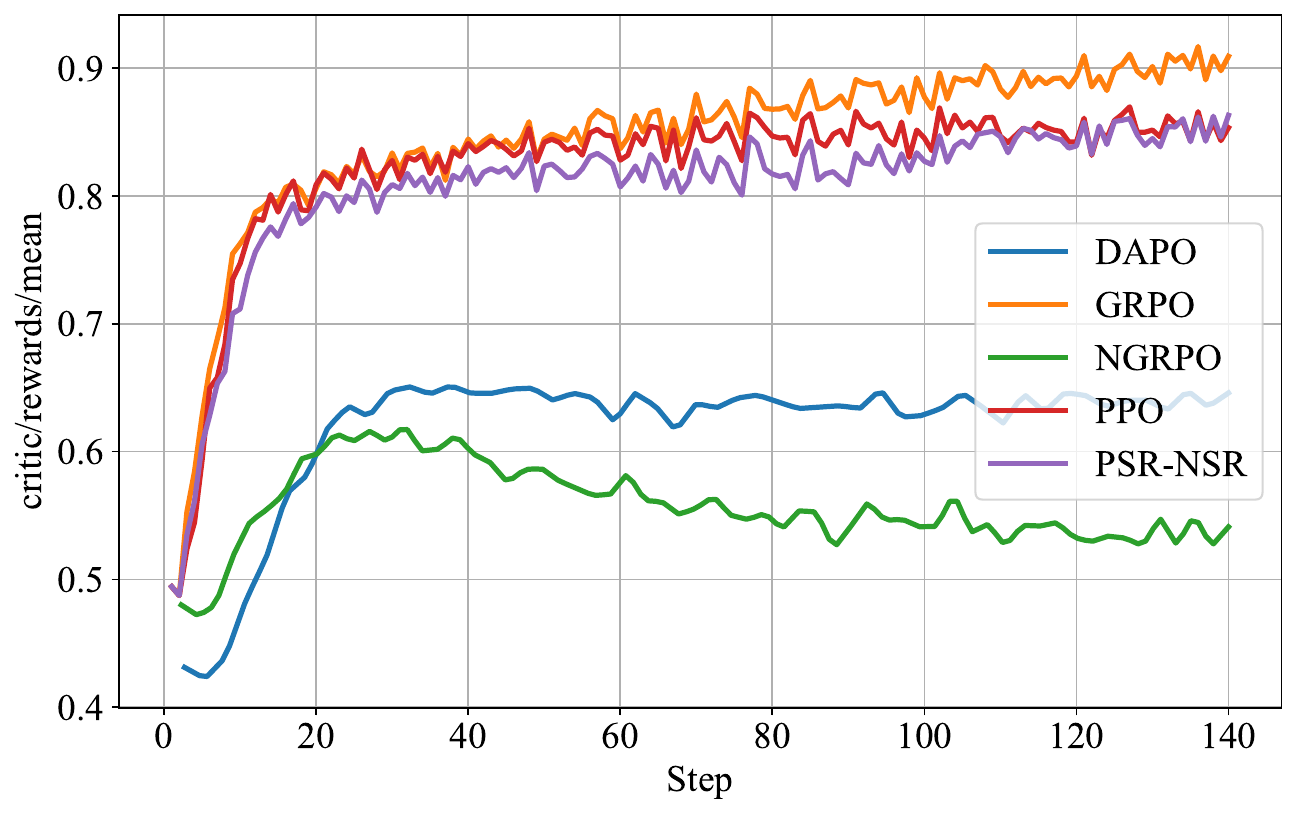}
    \caption{Reward}
    \end{subfigure}
    \hfill
    \begin{subfigure}[t]{0.48\textwidth}
    \centering
        \includegraphics[width=\linewidth]{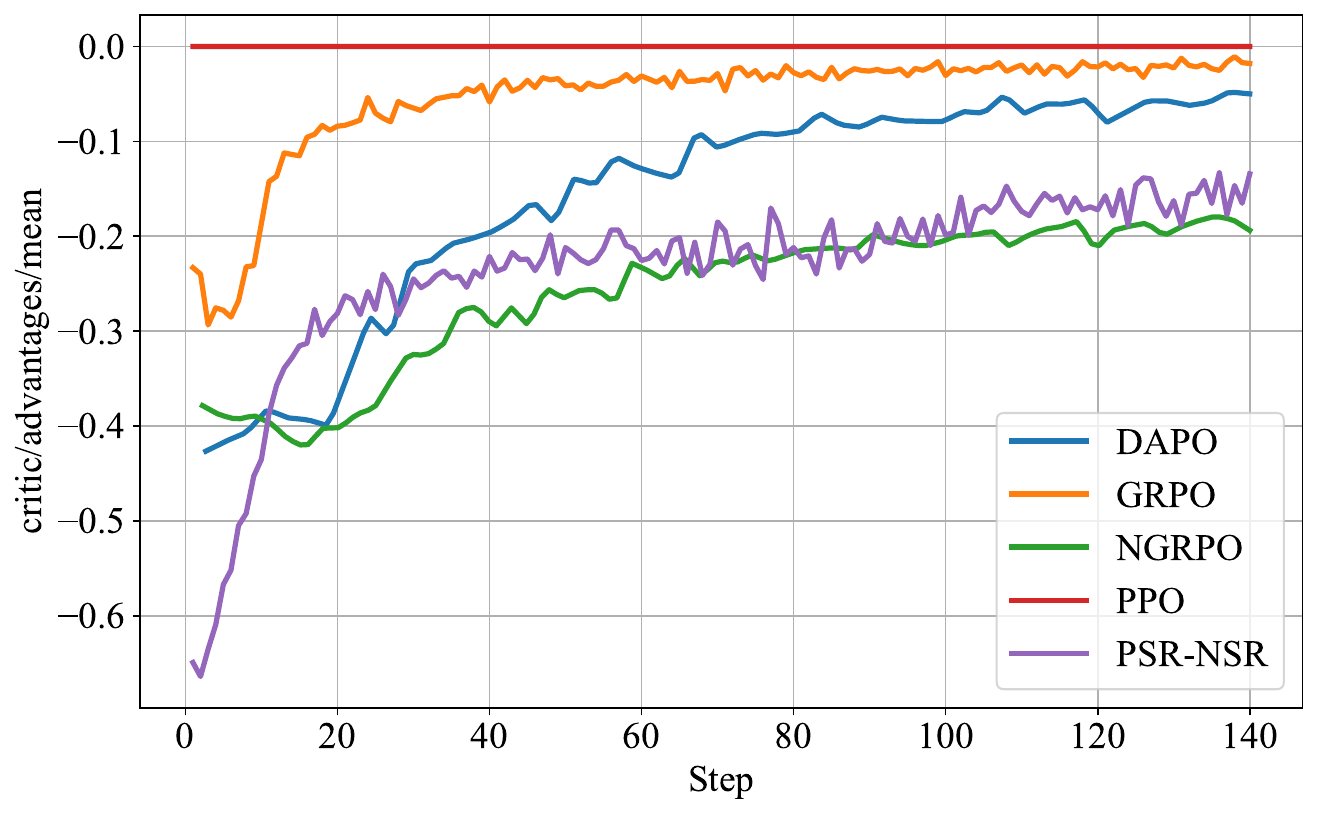}
    \caption{Advantage}
    \end{subfigure}
\caption{Reward and Advantage dynamics during training for different reinforcement learning methods.}
\label{fig:reward&advantage}
\end{center}
\end{figure}

Figure~\ref{fig:reward&advantage} shows that the lower average rewards for NGRPO and DAPO (left) are an expected artifact of their data filtering strategy, which removes high-reward, homogeneous-correct groups from the training data. The lower average advantage of NGRPO (right) is also an intended consequence of its design. The virtual $r_{\text{max}}$ sample systematically raises the reward baseline, creating a stronger negative signal to drive exploration. Similarly, PSR-NSR's lower advantage stems from its use of a fixed, un-normalized penalty for incorrect samples.

\subsection{Impact of Virtual Reward Magnitude and Count}
We conducted an ablation study to validate our design choices for the virtual reward mechanism, with results in Table~\ref{tab:ablation_reward}. The findings confirm that using the task's maximum reward ($r_{\text{max}}=1.0$) is optimal; a minimum reward impairs learning, while a medium reward provides insufficient exploratory pressure. The table also shows that adding more than one virtual reward leads to excessive and unproductive exploration, degrading performance. Therefore, a single virtual $r_{\text{max}}$ sample strikes the optimal balance between effective exploration and training stability.

\begin{table}[htbp]
\centering
\caption{Ablation studies on the magnitude and number of virtual rewards, evaluated by Pass@k AUC.}
\label{tab:ablation_reward}
\begin{tabular}{lc|lc}
\toprule
\multicolumn{2}{c|}{\textbf{Impact of Magnitude}} & \multicolumn{2}{c}{\textbf{Impact of Count}} \\
\textbf{Magnitude} & \textbf{AUC} & \textbf{Count} & \textbf{AUC} \\
\midrule
\multicolumn{4}{c}{\textbf{AIME2025}} \\
\midrule
max\_reward & \textbf{31.28} & 1 & \textbf{31.28} \\
min\_reward & 29.06 & 2 & 30.96 \\
medium\_reward & 30.39 & 4 & 28.85 \\
\midrule
\multicolumn{4}{c}{\textbf{AMC}} \\
\midrule
max\_reward & \textbf{86.09} & 1 & \textbf{86.09} \\
min\_reward & 84.29 & 2 & 84.79 \\
medium\_reward & 84.98 & 4 & 84.21 \\
\bottomrule
\end{tabular}
\end{table}

\end{document}